\documentclass{article}

\usepackage{float}
\usepackage{multirow}
\usepackage{PRIMEarxiv}
\usepackage{amsmath}
\usepackage[utf8]{inputenc} 
\usepackage[T1]{fontenc}    
\usepackage{hyperref}       
\usepackage{url}            
\usepackage{booktabs}       
\usepackage{amsfonts}       
\usepackage{nicefrac}       
\usepackage{microtype}      
\usepackage{lipsum}
\usepackage{fancyhdr}       
\usepackage{graphicx}       
\graphicspath{{media/}}     

\title{Automated Optimal Layout Generator for Animal Shelters: A framework based on Genetic Algorithm, TOPSIS and Graph Theory
}

\author{
  Arghavan Jalayer\\
  School of Architecture and Environmental Design, \\
  Iran University of Science and Technology\\
  Tehran, Tehran, Iran\\
  \texttt{arghavandjalayer@gmail.com} \\
   \And
  Masoud Jalayer\\
  Dipartimento di Ingegneria Gestionale, \\
  Politecnico di Milano\\
  Via Lambruschini 4/b, 20156, Milan, Italy\\
  \texttt{masoudjalayer@polimi.it} \\
   \And
  Mehdi Khakzand\\
  School of Architecture and Environmental Design, \\
  Iran University of Science and Technology\\
  Tehran, Tehran, Iran\\
   \And
  Mohsen Faizi\\
  School of Architecture and Environmental Design, \\
  Iran University of Science and Technology\\
  Tehran, Tehran, Iran\\
}

\begin{document}
\maketitle

\begin{abstract}
Overpopulation in animal shelters contributes to increased disease spread and higher expenses on animal healthcare, leading to fewer adoptions and more shelter deaths. Additionally, one of the greatest challenges that shelters face is the noise level in the dog kennel area, which is physically and physiologically hazardous for both animals and staff. This paper proposes a multi-criteria optimization framework to automatically design cage layouts that maximize shelter capacity, minimize tension in the dog kennel area by reducing the number of cages facing each other, and ensure accessibility for staff and visitors. The proposed framework uses a Genetic Algorithm (GA) to systematically generate and improve layouts. A novel graph theory-based algorithm is introduced to process solutions and calculate fitness values. Additionally, the Technique for Order of Preference by Similarity to Ideal Solution (TOPSIS) is used to rank and sort the layouts in each iteration. The graph-based algorithm calculates variables such as cage accessibility and shortest paths to access points. Furthermore, a heuristic algorithm is developed to calculate layout scores based on the number of cages facing each other. This framework provides animal shelter management with a flexible decision-support system that allows for different strategies by assigning various weights to the TOPSIS criteria. Results from cats' and dogs' kennel areas show that the proposed framework can suggest optimal layouts that respect different priorities within acceptable runtimes. 
\end{abstract}


\section{Introduction}
\label{sec:1}

Layout design is a critical decision-making phase in various fields, including plant design \cite{guo2019}, production planning \cite{zhang2019}, facility layout design \cite{masoud2019}, building design \cite{wang2016}, interior design \cite{matsuno2019}, solar farm spatial panel installation\cite{zhong2020}, and wind farm installation \cite{gualtieri2019}. In animal shelters, an effective cage layout design can enhance welfare by reducing noise levels \cite{coppola2010} and overcrowding \cite{doyle2020}. Several measures have been proposed in the literature to enhance the welfare of animal shelters, including accommodation design \cite{wagner2018}, capacity for care \cite{karsten2017}, environmental enrichment \cite{perry2020}, and social contact \cite{dudley2015}. However, optimizing cage layouts in animal shelters remains a gap in the literature.

\subsection{Computer-Aided Layout Design}

Layout design involves decisions regarding the optimal spatial configuration based on a set of constraints. This solution is highly dependent on the application and the specific requirements of the problem. Numerous computer-aided frameworks have been proposed to address layout design optimization in different industries.

Derhami et al. (\cite{derhami2020}) proposed a simulation-based algorithm to optimize block-stocking in a warehouse layout, considering aisles, cross-aisles, and bay depths to minimize material handling costs and maximize space utilization. Kang et al. (\cite{kang2018}) addressed facility layout problems using a Cuckoo Search algorithm and a random-key encoding scheme, finding near-optimal layouts for machinery cells in flexible manufacturing systems. Dino (\cite{dino2016}) developed an evolutionary framework named EASE for 3D architectural space layout, which includes sub-heuristics to construct valid spatial layouts and a mathematical model to check constraint satisfaction.

Genetic Algorithms (GA) have been successfully used in various layout design problems. Islier (\cite{islier1998}) proposed a GA for facility layout design that minimizes transportation load between departments, maximizes departmental compactness and aligns requested and available areas. Wu et al. (\cite{wu2020}) used a GA to design wind farms, optimizing for annual economic benefit considering energy, land, and cable costs. \cite{wu2007} developed a hierarchical GA for cellular manufacturing design, focusing on cell formation, group layout, and scheduling. Luo et al. (\cite{luo2019}) designed a GA for layout optimization in green logistic parks, aiming to maximize spatial utilization and minimize material handling costs. Kumar and Cheng (\cite{kumar2015}) employed a GA for site layout planning in congested construction sites, utilizing Building Information Management to reduce travel distances by 13\% compared to traditional methods.

Multi-criteria decision-making (MCDM) approaches have also been used to address layout planning, typically evaluating manually designed layouts. Sharma and Singhal (\cite{sharma2017}) proposed a fuzzy TOPSIS technique for facility layout planning, evaluating criteria such as initial data requirements and future expansion considerations. Murugesan et al. (\cite{murugesan2020}) used the Analytical Hierarchy Process (AHP) to improve the operational performance of an India National Sorting Hub, evaluating layouts based on personnel and material flow, throughput time, space consumption, safety, comfort, and noise control. Hervás-Peralta et al. (\cite{hervas2020}) addressed layout design for terminals and ports handling dangerous goods, using criteria like economic efficiency, capacity, and safety, evaluated by a panel of experts using AHP.

\subsection{Problems in Animal Shelters}
Animal shelters face distinct problems such as overpopulation, hunger, lack of space, high euthanasia rates, and seasonal fluctuations in animal intake \cite{olmez2019,karsten2017}. Optimal capacity for animal housing is crucial for successful shelter management, directly impacting animal health and well-being. North American shelters commonly house cats in barren, individual cages to reduce disease spread, despite drawbacks identified in both individual and group housing \cite{ottway2003, gourkow2006}. Factors such as expected length of stay (LOS), animal type, disease risk, and cost influence housing decisions \cite{wagner2018}. Inadequate housing capacity impedes shelters' ability to meet key goals, leading to high euthanasia rates \cite{karsten2017}.

High noise levels in shelters, often exceeding 119 dB, pose additional challenges \cite{coppola2010}. OSHA's permissible noise exposure limit requires that workers in environments with 115 dB noise levels be exposed for less than 15 minutes to avoid hearing loss risks \cite{osha2020}. High noise levels negatively impact both workers and animals, with dogs showing increased aggression and stress when cages face each other \cite{coppola2010}. Conversely, for felines, eye contact can reduce stress and anxiety, with cats at eye level more likely to be adopted \cite{janke2017}.

\subsection{Objectives of This Study}
This study aims to reduce noise levels in animal kennels and increase capacity through a multi-criteria framework for optimal cage layouts. The framework considers requested capacity, accessibility, and welfare, allowing shelter management to prioritize different criteria with flexible weight assignments. The objectives are:

Provide management with the flexibility to assign weights to prioritize different criteria.
Generate optimal layouts based on given weights, priorities, and configurations with sufficiently fast runtimes.
The remainder of the paper is organized as follows: Section~\ref{sec:2} presents the framework for multi-criteria layout optimization, detailing related materials and operators, criteria, and the graph-based algorithm used for calculation. Section~\ref{sec:3} introduces the configuration of animal shelters and cage requirements and depicts proposed solutions for different scenarios. Finally, conclusions and opportunities for future studies are summarized in Section~\ref{sec:4}.

\section{The Proposed Framework and Materials}
\label{sec:2}

This section presents the details of the proposed model designed to automatically arrange cage layouts, optimizing space usage in animal shelters. The space usage optimizer algorithm considers given space limitations and optimizes multiple criteria simultaneously. The criteria are as follows:
\begin{itemize}
  \item Number of Accessible Cages (AC): The cages should be arranged in a layout that meets standard requirements, providing maximum animal capacity.

  \item Number of Inaccessible Cages (IC): The cages should be arranged so that no cage resides in an inaccessible location.

  \item The Longest Shortest-Path (LSP) and the Average of the Shortest-Paths (ASP): To facilitate convenient movement within the shelter, this paper proposes a graph-based shortest path solution, detailed in Section 2.2.4.

  \item The Number of Cages Facing Each Other (CF): As mentioned in Section 1, dogs' cages facing each other increase aggression and barking. Thus, reducing the number of cages facing each other is crucial for minimizing shelter noise.
\end{itemize}

To optimize these criteria simultaneously, we employ a Multi-Criteria Decision-Making (MCDM) solver technique, as described in Section 2.1.

\begin{figure}[H]
    \centering
    \includegraphics[width=10cm]{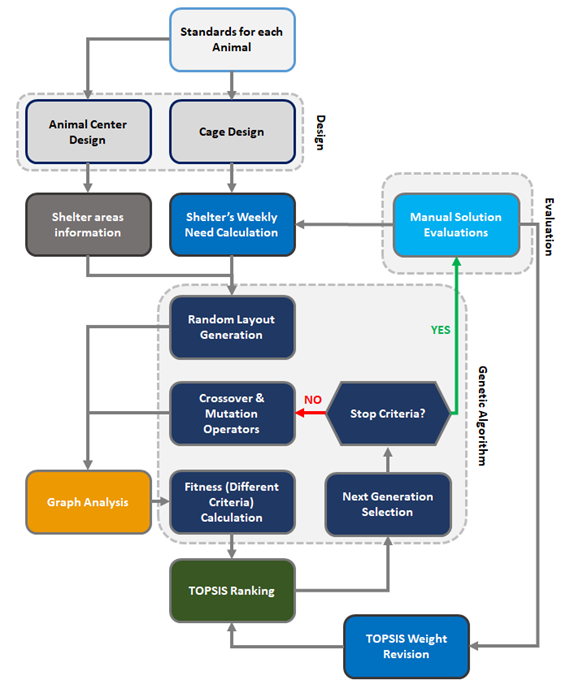}
    \caption{The steps of the proposed framework.}
    \label{fig:1}
\end{figure}

The steps of the proposed animal shelter layout optimization framework are illustrated in Figure~\ref{fig:1}.

\subsection{TOPSIS Algorithm}
Among the various Multi-Criteria Decision-Making (MCDM) techniques, the "Technique for Order of Preference by Similarity to Ideal Solution" (TOPSIS) is one of the most popular. Introduced in 1981 by Hwang and Yoon, TOPSIS provides numerical solutions for MCDM problems, and it is preferred here due to its fewer rank reversals compared to other techniques \cite{afful2015fuzzy}.

MCDM techniques are employed in various research fields, such as building energy management, composite chemistry, agriculture, and financial management. Recently, two related papers have utilized TOPSIS for designing building construction layouts.

TOPSIS consists of seven steps, which are applied to our layout design problem as follows:

\begin{enumerate}
    \item Generating the Decision Matrix $X$:

\begin{equation}
    X=\begin{bmatrix}
        x_{11} & \cdots & x_{1n} \\
    \vdots & \ddots & \vdots \\
    x_{m1} & \cdots & x_{mn}
    \end{bmatrix}
\end{equation}

    \item Normalizing the Decision Matrix $X \rightarrow R$:

\begin{equation}
    r_{ij} = \frac{x_{ij}}{\sqrt{\sum_{i=1}^{m} x_{ij}^2}} \quad \forall j
\end{equation}

    \item Determining the weighted normalized matrix $[t_{ij}]_{m \times n} = [w_{j} \cdot r_{ij}]_{m \times n}$ representing the relative importance of each attribute.

    \item Calculating the Positive and Negative Ideal Solutions:
    \begin{equation}
        t_j^+ = \left\{
\begin{array}{ll}
\max t_{ij} \quad & \text{if } j \in J^+ \\
\min t_{ij} \quad & \text{if } j \in J^-
\end{array}
\right., \quad
 t_j^- = \left\{
\begin{array}{ll}
\max t_{ij} \quad & \text{if } j \in J^- \\
\min t_{ij} \quad & \text{if } j \in J^+
\end{array}
\right.
    \end{equation}

    \item Distance to Positive and Negative Ideal Solutions:
    \begin{equation}
        S_i^+ = \sqrt{\sum_{j=1}^{n} (t_{ij} - t_j^+)^2}, \quad
        S_i^- = \sqrt{\sum_{j=1}^{n} (t_{ij} - t_j^-)^2}
    \end{equation}
    \item Relative Closeness to the Ideal Solution:
$ C_i^* = \frac{S_i^-}{S_i^- + S_i^+}$

    \item Sorting the Proposed Layouts by $C_i^*$ Values in Descending Order
\end{enumerate}

\subsection{Genetic Algorithm (GA)}

Introduced by Holland in 1974, Genetic Algorithms (GA) are evolutionary computational approaches inspired by natural selection processes. GA concepts include chromosomes, genes, mutation, and crossover offspring, designed to address optimization problems where deterministic approaches are infeasible. GA is a stochastic method, meaning its results rely on random or pseudo-random variables and generators, and it is considered a heuristic due to the lack of rigorous theoretical justification.

GA works by treating each solution as a chromosome, containing a set of genes representing the solution's parameters. It involves altering genes and combining chromosomes to create new generations iteratively. A generation is ranked by fitness values, and selected individuals generate offspring for the next generation through mutation and crossover operations.

\subsubsection{Creating the Initial Population}
Given the complexity of the problem, regulations are defined to shrink the search space. A "resolution" variable 
$r$ converts the shelter area $m \times n$ square meters into matrix $A$:

\begin{equation}
    A = (a_{ij}) \in \mathbb{N}^{x,y}; \quad x = \left\lceil \frac{m}{r} \right\rceil, \quad y = \left\lceil \frac{n}{r} \right\rceil
\end{equation}

Cages can only have four orientations: upward, downward, rightward, and leftward. The number of possible placements for a cage of length 
$l$ and width $w$ in an area with no obstacles can be calculated as:

\begin{equation}
    n_{\text{poss}} = 2(m - c_{\text{len}} - \left\lceil \frac{g}{r} \right\rceil + 1)(n - c_{\text{wid}} + 1) + 2(m - c_{\text{wid}} + 1)(n - c_{\text{len}} - \left\lceil \frac{g}{r} \right\rceil + 1)
\end{equation}
where $c_{\text{wid}} = \frac{w}{r}, \quad c_{\text{len}} = \frac{l}{r} $ and 
$g$ is the required free space in front of the cage’s door. For a single cage with $l=1.5$ and $w=0.75m$ in a $20m x 20m$ area, and $g=2.5m$, there are 5148 possibilities with $r=0.5m$.

\subsubsection{Placement Strategies}
The algorithm randomly generates the layouts, considering the space requirements to avoid obstructing the doors and the lanes. To this aim, an algorithmic loop is created to place the cages one by one and employs one of the pre-defined strategies to add a new cage to the area. The strategies to place a new cage are described at Table~\ref{tab:1}.

\begin{table}[h]
\caption{Different strategies to place a new cage}
\label{tab:1}
\begin{tabular}{p{3cm}p{8.5cm}p{4cm}}
\toprule
Strategy name    & Description                                                                                                                                                                               & Example \\
\midrule
Total Randomness & It randomly places the new cage in a place with a random orientation while it complies with the space requirements.                                                                          & -       \\
Confrontation    & An existing cage in front of which there is a vacant place for a new cage is selected. The algorithm places the new cage in front of the selected cage with the opposite orientation. &    \includegraphics[]{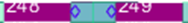}     \\
Neighbourhood & An existing cage which has sufficient vacant space in its adjacency is selected. The algorithm places the new cage beside the selected cage with the same orientation.              &   \includegraphics[]{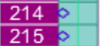}      \\
Back-to-back     & An existing cage which has sufficient vacant space in its behind is selected. The algorithm places the new cage behind the selected cage with an opposite orientation.              &     \includegraphics[]{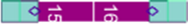}    \\
Aligned          & An existing cage which has sufficient vacant space in its behind is selected. The algorithm places the new cage behind the selected cage with the same orientation.                 &    \includegraphics[]{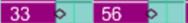}    \\
\bottomrule
\end{tabular}
\end{table}

\begin{figure}[H]
    \centering
    \includegraphics[width=\linewidth]{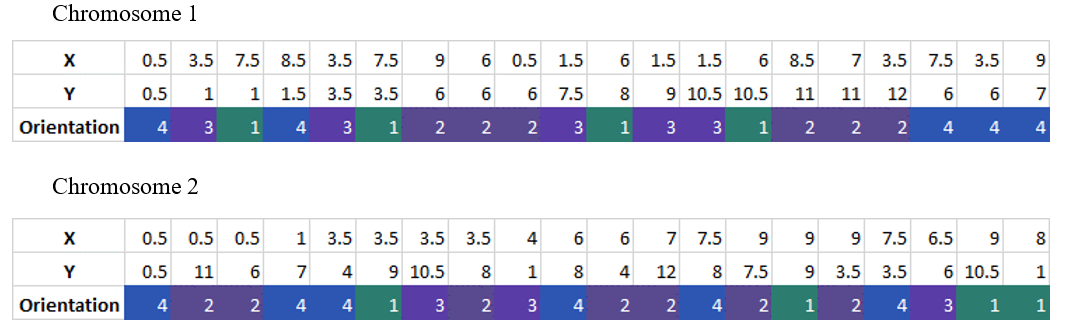}
    \caption{Two initial chromosomes for a shelter with 20 cages.}
    \label{fig:2}
\end{figure}

\begin{figure}[H]
    \centering
    \includegraphics[width=\linewidth]{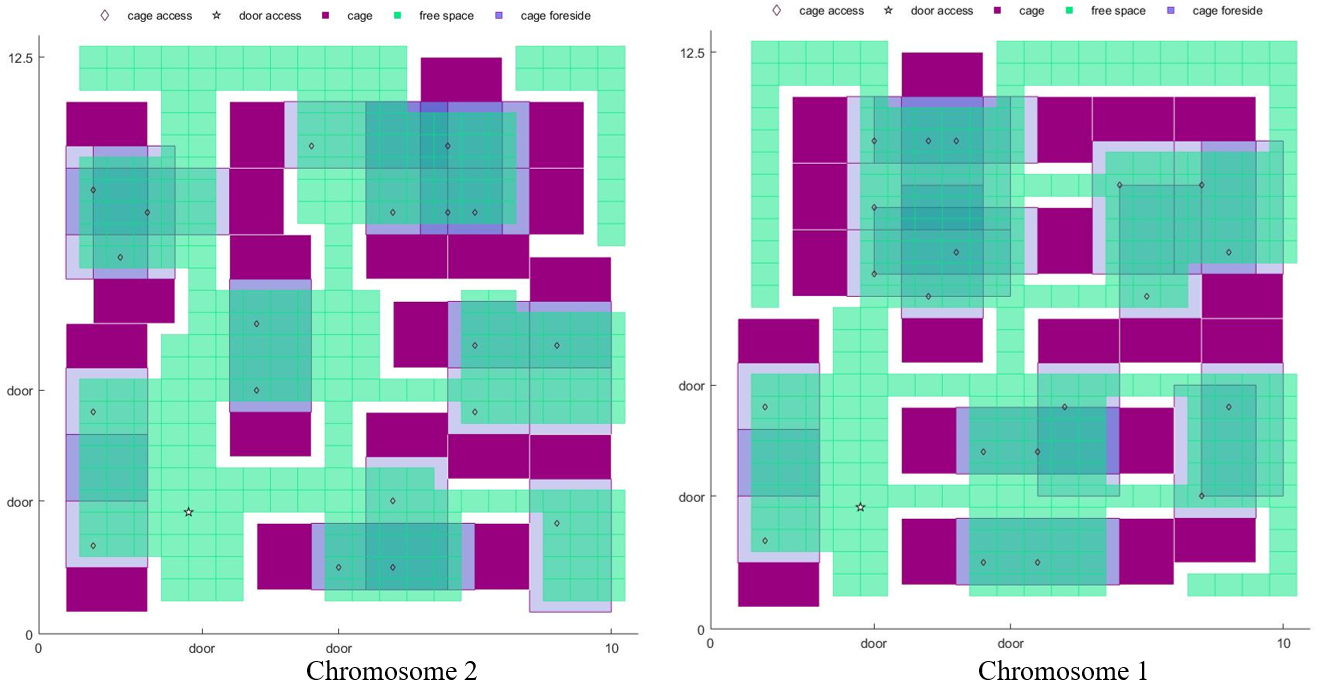}
    \caption{The corresponding layouts of the two initial chromosomes.}
    \label{fig:3}
\end{figure}

The algorithm stochastically uses these five strategies to fill the shelter with the desired number of cages. Figure~\ref{fig:2} illustrates two different initial chromosomes which consist of three string genes, $x$, $y$ and orientation for a shelter area with 20 cages. Each column of the chromosomes represents a cage’s placement. The first two arrays show the placement of the cage’s lower left corner in matrix $A$, and the third array represents the orientation of the cage in the shelter. Figure~\ref{fig:3} depicts the corresponding layouts of these two chromosomes in the shelter with dimensions of $10\times 12.5$ square meters with three entrances.

\subsubsection{Crossover and Mutation Operators}
\begin{figure}[h]
    \centering
    \includegraphics[width=13cm]{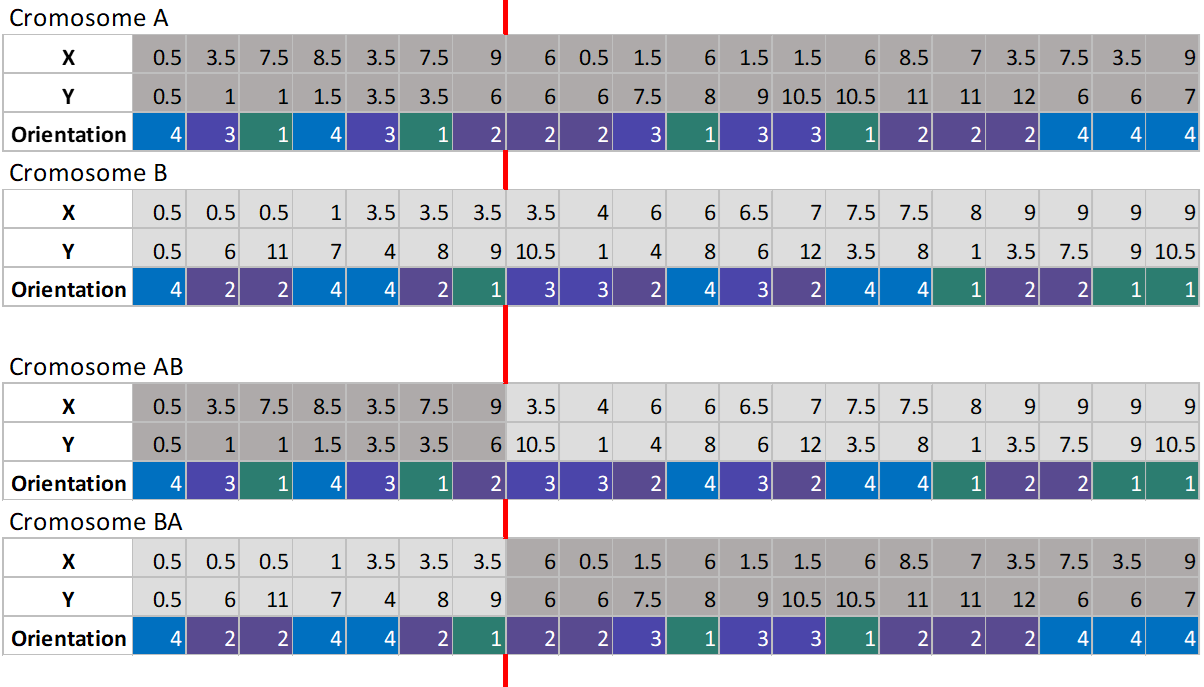}
    \caption{The crossover operation to create chromosomes AB and BA from parents A and B.}
    \label{fig:4}
\end{figure}

\textbf{Crossover Operator}: Designing the crossover operator is usually the hardest part of developing a GA. Programming this operator is highly dependent on the problem's features and constraints and so often is done heuristically(Poon and Carter 1995). Re
After creating the initial population, with the aforementioned steps. In this problem, each pair of crossover offspring is made of a pair of randomly selected parents. In order to select the parents, the paper uses a roulette wheel selection algorithm. Then, one of the two axes will be randomly chosen to select a cutting point where each parent will be split in two. Consequently, by swapping the parts, new offspring are created, inheriting parts of their parent’s genes. Though, such exchanges can result in nonpractical solutions since there might be cages in the newborn chromosomes where the requirements are not passed or there are cages overlapping each other. Our aim is to design a meticulous encoding schema and operators which can safeguard the feasibility of the solutions; Thus, the algorithm checks the cages, one by one. If there is any conflict, the corresponding algorithm wipes out the cage and places a new cage in the layout using one of the five placement strategies, randomly. Moreover, if there are fewer than required cages in the newborn chromosome, the algorithm places new cages until it is satisfied. Figure~\ref{fig:4} illustrates how two chromosomes A and B, share their genes to make two crossover offspring.

\begin{figure}[h]
    \centering
    \includegraphics[width=13cm]{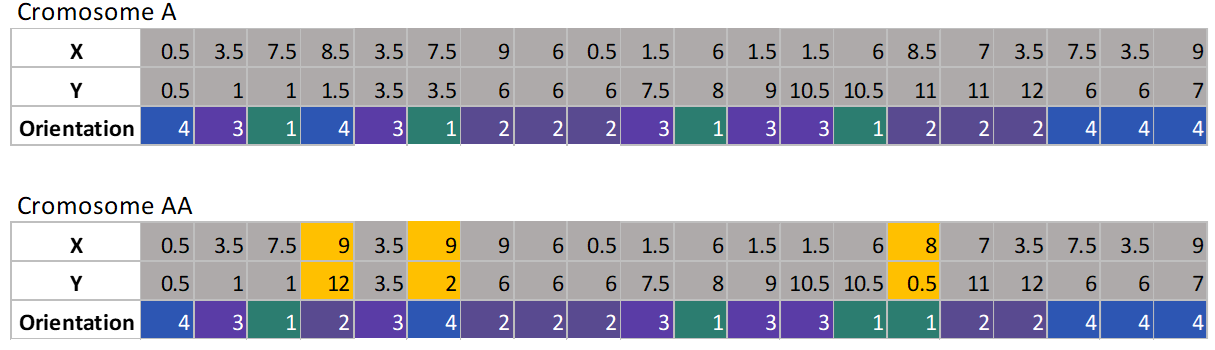}
    \caption{The mutation operation to create chromosomes AA from parent A.}
    \label{fig:5}
\end{figure}

\textbf{Mutation Operator}: Without a mutation operator GA tends to get stuck in a local optimum point and misses the opportunity to have a wider search within the solution space (Hesser and Männer 1990). Just like the crossover operator, the mutation operator chooses its parents with a roulette wheel selection algorithm. For each mutation offspring, unlike the crossover, the operator only needs one parent. During the mutation process, the newborn offspring inherits a random number of genes from the parent’s chromosome. Next, the algorithm places the cages one by one, using five placement strategies until the offspring’s chromosome gets complete (see Figure~\ref{fig:5}).

\subsubsection{Fitness Operator}
The GA evaluates each chromosome's quality based on four criteria: Number of Accessible Cages (AC), Longest shortest path (LSP) to the entrance, Average of Shortest-Paths (ASP) to the entrance, and Number of Cages Facing Each Other (CF).

Shortest Paths Calculation
As mentioned earlier, the convenience of access to the cages is one of the attributes that determine the fitness of a layout. This study assumes that the ideal solution has the lowest average value of distances between the cages and the entrances. It also assumes that the ideal solution has the lowest distance of the furthest cage to the entrance, which is equal to the maximum of the shortest paths in each solution. This paper proposes a heuristic approach to calculate the distance of a cage to the entrance.

For each solution, we first consider the binary matrix of the area $A$ (Equation 5), containing all movement obstacles such as walls, columns, and cages. An array $a_{ij}$
takes the value 1 only if there is no obstacle within a radius of 0.5 meters. We consider a graph 
$G$ composed of $x\times y$ nodes:

\begin{equation}
G = (V, E); \quad V = \{ v_{1,1}, \ldots, v_{x,y} \}
\end{equation}

where 
$V$ represents the set of nodes. If there are two adjacent arrays with values equal to one, there is an edge between the corresponding nodes with a weight of 1. The edge 
$e_{ij, kl}$ is calculated as follows:

\begin{equation}
    e_{ij, kl} = \left\{
\begin{array}{ll}
1 & \text{if } a_{ij} = a_{kl} = 0 \\
0 & \text{otherwise}
\end{array}
\right. \text{for} \quad |i-k| \times |j-l| = 0 \quad \text{and} \quad |i+j-k-l| = 1
\end{equation}

Accordingly, there will be a network of connections between adjacent points in the area where staff can move without any inconvenience. To find the shortest path between two points in the area, the proposed framework uses the built-in shortest-path algorithm in MATLAB R2019a.

\textbf{\textit{Calculating Different Criteria-}}
As mentioned above, some of the newborn chromosomes may contain cages whose access paths are blocked by other cages or obstacles. To find the number of accessible cages, the algorithm calculates the shortest path to the entrances. If a cage has an empty solution, it means that it’s an inaccessible cage.

\begin{figure}[h]
    \centering
    \includegraphics[width=13cm]{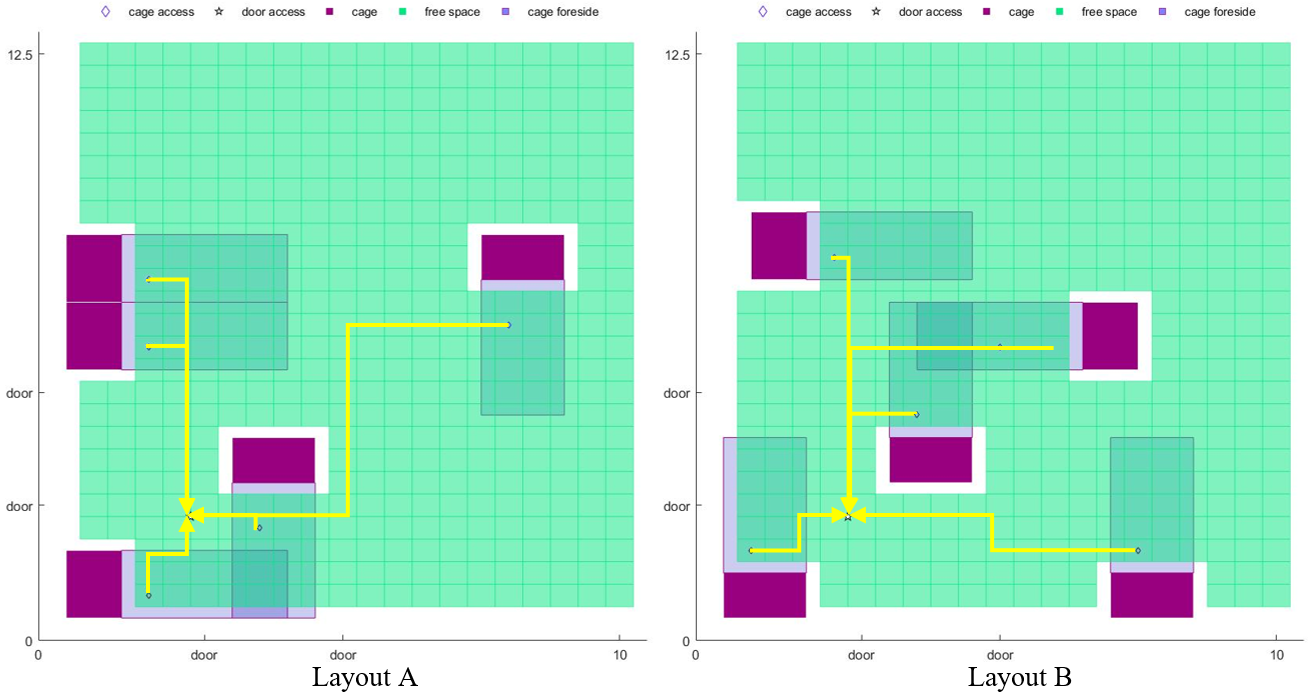}
    \caption{Two different layouts with the same ASP values.}
    \label{fig:6}
\end{figure}

\begin{figure}[h]
    \centering
    \includegraphics[width=13cm]{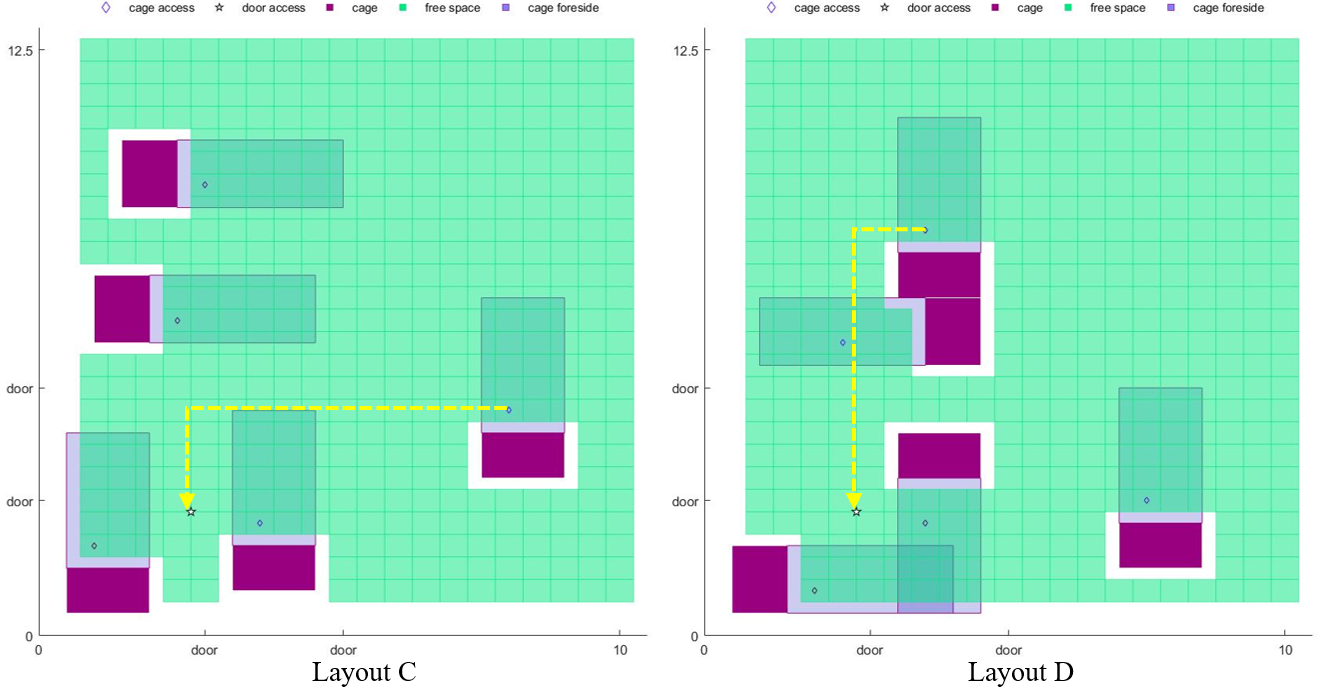}
    \caption{Two different layouts with the same LSP values.}
    \label{fig:7}
\end{figure}

The Longest Shortest-Path (LSP) and the Average of Shortest-Paths (ASP) are the other two criteria that can be calculated using the shortest-path algorithm.
Figure~\ref{fig:6} and \ref{fig:7} illustrate two pairs of layouts with the same ASP and LSP values.

\textbf{\textit{Calculating CF Values-}}
To calculate the CF values, the algorithm takes each cage, called the "first cage," and draws a vector from the first cage’s access point to all other cages’ access points with different orientations. Vectors not interrupted by any obstacle, including cage bodies, are considered eye-contact vectors. Therefore, the CF value of a layout is calculated as:

\begin{equation}
    s_{ij} = \left\{
\begin{array}{ll}
0 & \text{if there is an obstacle between } i \text{ and } j \\
1 & \text{otherwise}
\end{array}
\right.
\end{equation}

\begin{equation}
    CF = \sum_{j=1}^N \sum_{i=1, i \neq j}^N \frac{s_{ij}}{l_{ij}}
\end{equation}

where $l_{ij}$ denotes the length of an eye-contact vector between cages $i$ and $j$. $N$ represents the number of cages

\section{Results and Discussion}
\label{sec:3}
\subsection{Specifications of the Case}
\begin{figure}[h]
    \centering
    \includegraphics[width=15cm]{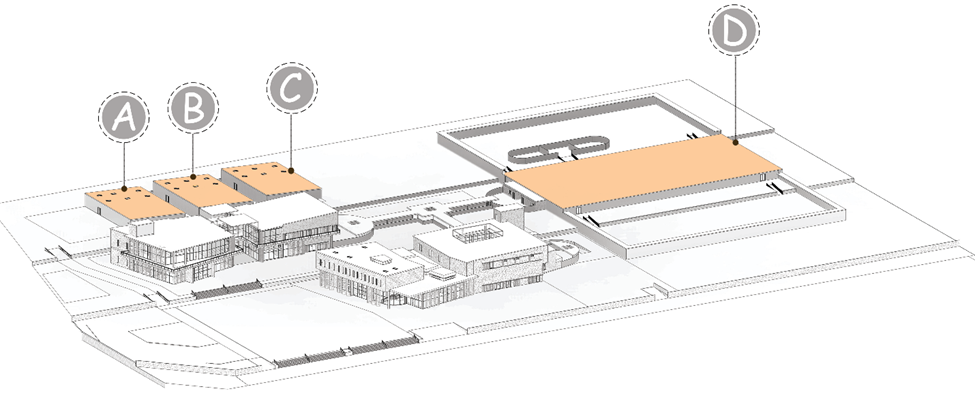}
    \caption{The perspective of the animal care center.}
    \label{fig:8}
\end{figure}

To demonstrate the practicality of the framework, the model was implemented on an animal care center planned to be located on Kharrazi Expressway, Tehran, Iran. The complex consists of four separate rectangular shelters: three for cats and a more spacious one for dogs, located approximately 50 meters apart from the cat shelters. Figure~\ref{fig:8} illustrates a perspective view of the animal care center, and Table~\ref{tab:2} provides the specifications of the shelters.

\begin{table}[h!]
\caption{Specifications of the shelters}
\label{tab:2}
\centering
\begin{tabular}{ccccc}
\hline
\textbf{ID} & \textbf{Animal} & \textbf{Length} & \textbf{Width} & \textbf{\# Doors} \\
\hline
A & Cats & 25 m & 20 m & 2 \\
B & Cats & 30 m & 20 m & 3 \\
C & Cats & 30 m & 20 m & 3 \\
D & Dogs & 81.2 m & 36.4 m & 6 \\
\hline
\end{tabular}
\end{table}

According to the HSUS guidelines (The Humane Society of the United States, 1999), an individual dog cage should have a kennel section of 1.2m by 1.8m and a run section of 1.2m by 2.4m, allowing the dog to engage in daily training activities. Therefore, this paper considers dog cages with dimensions of $1.4m$ by $4.2m$, compatible with a resolution of $r=1.4m$. For cats, the HSUS guidelines suggest individual cages of $0.91m$ by $0.91m$. For simplicity, this paper uses a resolution of $r=1.0m$ for the cat shelters.

\subsection{Optimization Results}
For these optimizations, this study employs a personal laptop with an Intel® i7 processor, 8GB of memory, and a Windows 11 operating system. The framework is fully coded in MATLAB R2020a.

\begin{figure}[h]
    \centering
    \includegraphics[width=15cm]{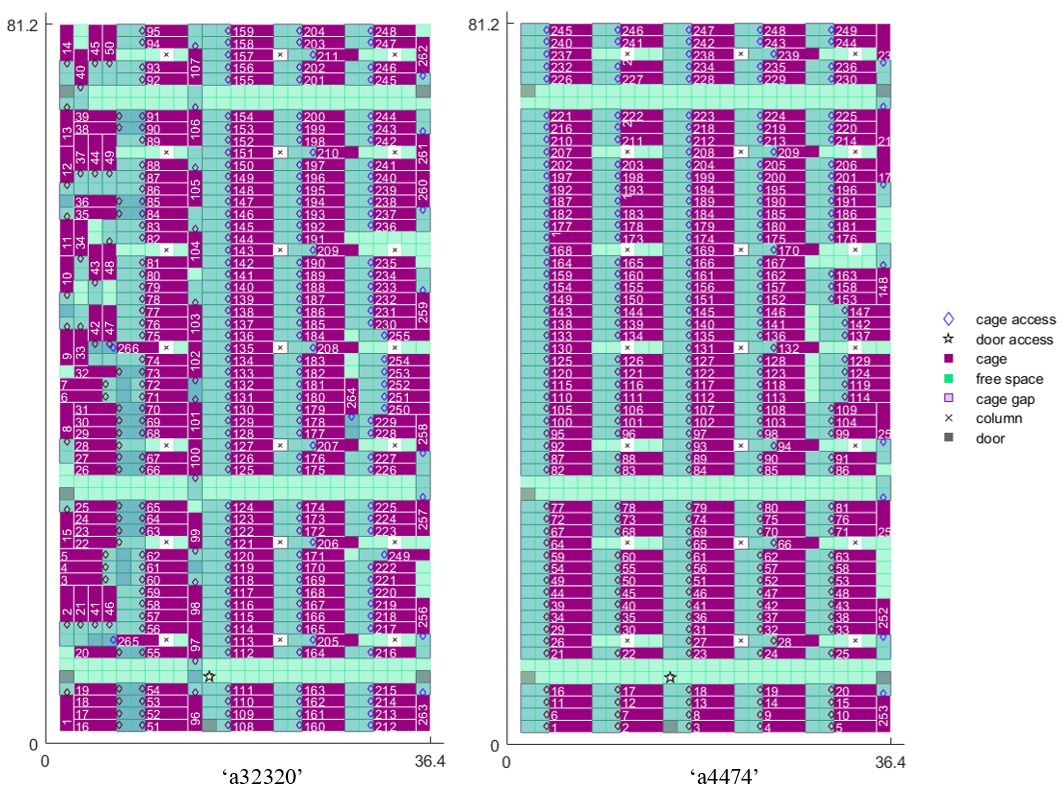}
    \caption{Near-optimum solutions found for shelter D for the first scenario (right), and the second scenario (left).}
    \label{fig:9}
\end{figure}

The optimization process involves generating and evaluating various cage layout configurations to find the optimal arrangement that maximizes space utilization, minimizes noise levels, and ensures accessibility. The results demonstrate the effectiveness of the proposed framework in designing practical and efficient layouts for animal shelters. The optimized layouts respect the constraints and guidelines, providing improved living conditions for the animals and facilitating easier management for the staff.

To optimize the cage layouts of shelter D, a combination of strategies—“randomness,” “neighborhood,” “back-to-back,” and “aligned”—were used with weights of 5\%, 35\%, 30\%, and 30\%, respectively. The “opponent” strategy was not used since it is less preferable for management to have dogs facing each other. It should be noted that the other strategies can still result in a new cage facing a pre-located cage.

For these GA searches, after a grid search, the population size was set to 24, with an equal share of 12 crossover and mutation children. In each iteration, the better 50\% of the population were kept for the next generation, while the rest of the chromosomes were wiped out.

The paper defines two scenarios for the management system to follow:
\begin{itemize}
    \item \textbf{\textit{Scenario I}}: The management aims to maintain the welfare of the shelter, opting for fewer dogs to achieve higher welfare.
    \item \textbf{\textit{Scenario II}}: The management tries to admit more dogs, prioritizing capacity.
\end{itemize}

For Scenario I, the weights were set to +44\% for AC, -4\% for LSP and ASP, -44\% for CF, and -4\% for minimizing the number of inaccessible cages (IC). The search configuration emphasized finding a layout with lower CF and higher AC, while assigning less priority to the other attributes. As shown in Table~\ref{tab:3}, chromosome ‘a4474’ had the best fitness among the population in the 40th iteration and was selected as the near-optimum solution. The final ranking shows that only a few confrontations occur in the near-optimal solution, ‘a4474,’ which managed to place 253 accessible cages in the shelter without any inaccessible ones.

Table~\ref{tab:4} shows the results of iteration 50 for Scenario II, where the main aim was to find a layout with higher AC importance. Consequently, ‘a32320’ had a high CF value of about 710, indicating many dogs could see each other, but it was still selected as the best layout. This compromise allowed for a higher number of dogs placed inside the kennels. Under these conditions, ‘a4474,’ which ranked first in the previous search, could not achieve better than the 6th place in the TOPSIS table. The proposed solutions for these two scenarios are depicted in Figure~\ref{fig:9}.

\begin{table}[H]
\caption{TOPSIS Scores for the Dogs’ Shelter in Iteration Number 50, Scenario I}
\label{tab:3}
\centering
\begin{tabular}{ccccccc}

\toprule
\multirow{2}{*}{ID \textbackslash Weight} & AC   & LSP  & ASP    & CF     & IC   & \multirow{2}{*}{Score} \\ \cline{2-6}
                                          & 44\% & -4\% & -4\%   & -44\%   & -4\% &                        \\
\midrule
a4474 & 253 & 66 & 32.609 & 0.4511 & 0 & 1 \\
a04184 & 250 & 65 & 32.564 & 20.721 & 7 & 0.9844 \\
a03297 & 246 & 65 & 32.325 & 0 & 0 & 0.9837 \\
a05265 & 203 & 64 & 31.158 & 0 & 19 & 0.8883 \\
a02721 & 203 & 64 & 31.158 & 0 & 19 & 0.8883 \\
a02533 & 164 & 64 & 32.421 & 0 & 30 & 0.8163 \\
a04137 & 164 & 64 & 32.421 & 0 & 30 & 0.8163 \\
a05642 & 164 & 64 & 32.421 & 0 & 30 & 0.8163 \\
a03338 & 220 & 64 & 30.568 & 226.18 & 38 & 0.7455 \\
a04295 & 241 & 67 & 30.278 & 332.88 & 23 & 0.6507 \\
a28928 & 252 & 81 & 38.393 & 372.71 & 15 & 0.6152 \\
a03144 & 148 & 72 & 34.453 & 379.8 & 60 & 0.5268 \\
a03134 & 184 & 71 & 34.011 & 454.66 & 60 & 0.4740 \\
a02756 & 206 & 67 & 33.238 & 511.46 & 36 & 0.4328 \\
a04189 & 209 & 65 & 32.746 & 528.95 & 37 & 0.4163 \\
a05214 & 207 & 65 & 32.488 & 584.17 & 37 & 0.3567 \\
a11768 & 263 & 65 & 31.954 & 700.63 & 0 & 0.3248 \\
a32320 & 266 & 76 & 34.808 & 710.4 & 0 & 0.3218 \\
a29798 & 259 & 65 & 31.946 & 696.74 & 4 & 0.3213 \\
a1461 & 259 & 65 & 31.946 & 696.74 & 4 & 0.3213 \\
a04798 & 260 & 76 & 35.831 & 739.19 & 5 & 0.2941 \\
a04752 & 257 & 76 & 35.638 & 785.49 & 4 & 0.2652 \\
a05447 & 239 & 67 & 33.669 & 748.3 & 26 & 0.2516 \\
a02699 & 235 & 64 & 32.157 & 871.17 & 26 & 0.1926 \\
\bottomrule
\end{tabular}
\end{table}

\begin{table}[H]
\centering
\caption{TOPSIS Scores for the Dogs’ Shelter in Iteration Number 50, Scenario II}
\label{tab:4}

\begin{tabular}{ccccccc}

\toprule
\multirow{2}{*}{ID \textbackslash Weight} & AC   & LSP  & ASP    & CF     & IC   & \multirow{2}{*}{Score} \\ \cline{2-6}
                                          & 90\% & -2\% & -2\%   & -3\%   & -3\% &                        \\
\midrule
a32320                                    & 266  & 76   & 34.808 & 710.4  & 0    & 1                      \\
a11768                                    & 263  & 65   & 31.954 & 700.63 & 0    & 0.99331                \\
a36133                                    & 258  & 76   & 34.841 & 536.4  & 3    & 0.98082                \\
a29798                                    & 259  & 65   & 31.946 & 696.74 & 4    & 0.96772                \\
a1461                                     & 259  & 65   & 31.946 & 696.74 & 4    & 0.96772                \\
a4474                                     & 253  & 66   & 32.609 & 0.4511 & 0    & 0.95701                \\
a40942                                    & 253  & 70   & 32.996 & 4.124  & 5    & 0.956                  \\
a28928                                    & 252  & 83   & 39.183 & 403.16 & 12   & 0.92061                \\
a22612                                    & 245  & 70   & 33.735 & 190.53 & 16   & 0.83716                \\
a26730                                    & 238  & 65   & 33.038 & 679.52 & 24   & 0.71828                \\
a17587                                    & 233  & 66   & 32.438 & 75.604 & 23   & 0.67173                \\
a32612                                    & 233  & 76   & 34.142 & 190.53 & 26   & 0.66719                \\
a2379                                     & 232  & 69   & 30.897 & 715.74 & 20   & 0.63884                \\
a33201                                    & 229  & 76   & 36.856 & 401.57 & 27   & 0.60529                \\
a3960                                     & 228  & 76   & 35.478 & 243.53 & 18   & 0.60033                \\
a37430                                    & 222  & 76   & 37.306 & 368.96 & 31   & 0.50824                \\
a35480                                    & 216  & 68   & 34.403 & 87.851 & 34   & 0.43616                \\
a27496                                    & 208  & 65   & 33.255 & 187.13 & 44   & 0.32134                \\
a25805                                    & 207  & 67   & 33.976 & 135.77 & 45   & 0.31099                \\
a20930                                    & 206  & 68   & 34.578 & 114.22 & 46   & 0.29922                \\
a7190                                     & 204  & 68   & 34.853 & 0      & 48   & 0.28288                \\
a14146                                    & 200  & 67   & 34.89  & 274.36 & 55   & 0.20798                \\
a12375                                    & 191  & 67   & 33.188 & 270.91 & 62   & 0.1116                 \\
a15585                                    & 187  & 77   & 35.497 & 207.22 & 66   & 0.10329    \\
\bottomrule
\end{tabular}
\end{table}

\begin{figure}[h]
    \centering
    \includegraphics[width=10cm]{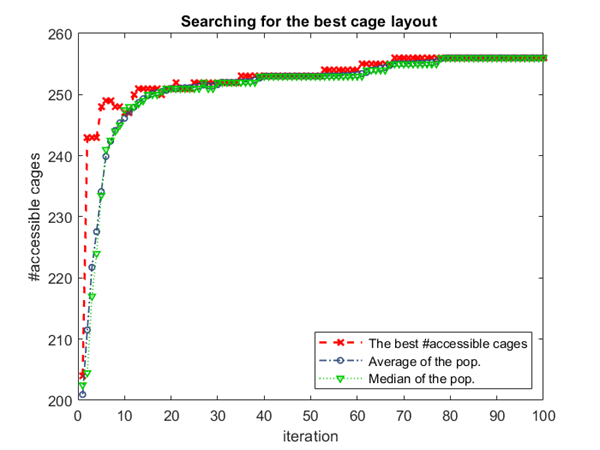}
    \caption{The GA performance at each iteration.}
    \label{fig:10}
\end{figure}

Figure~\ref{fig:10} illustrates the performance of the GA during the search for the best layout of Shelter D in the second scenario. The best, average, and median AC values for each iteration are depicted. It can be inferred that the population converges rapidly. With the mentioned configuration, the optimization runtime takes approximately 2 hours for each search.

\begin{figure}[h]
    \centering
    \includegraphics[width=10cm]{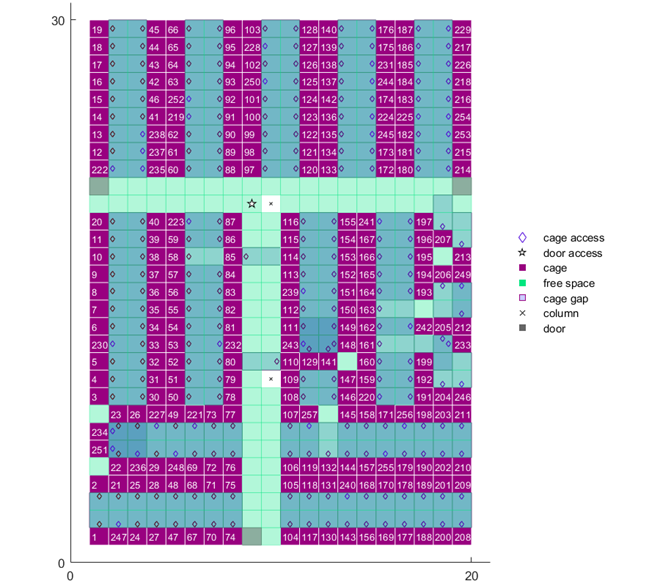}
    \caption{A near-optimum solution found for shelter B and C.}
    \label{fig:11}
\end{figure}

\begin{figure}[h]
    \centering
    \includegraphics[width=10cm]{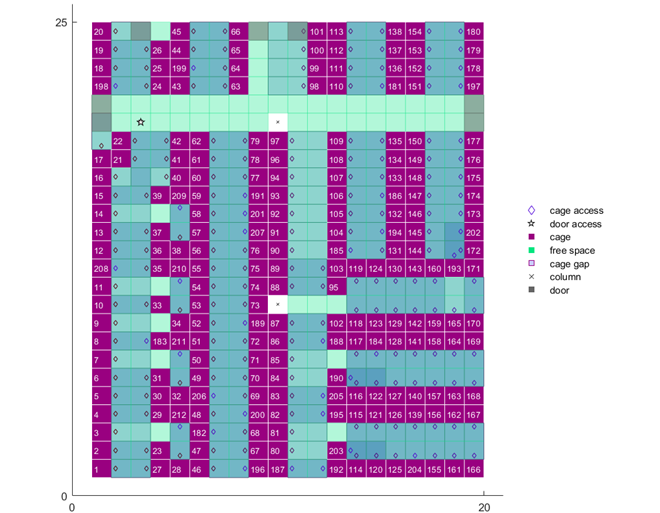}
    \caption{A near-optimum solution found for shelter A.}
    \label{fig:12}
\end{figure}

As previously mentioned, the confrontation of cats' cages does not negatively impact their welfare. Thus, to find the best layout, the “opponent” strategy is also implemented alongside other strategies for placing new cages in the three cat shelters (A, B, and C). Figures~\ref{fig:11} and \ref{fig:12} show the optimum solutions for the different cat shelters. According to the solutions, at each cage level, the animal care center can accommodate 257 accessible cages (AC) for the more spacious shelters B and C, and 212 AC for shelter A. The runtime performance information is provided in Table~\ref{tab:5}.

\begin{table}[h!]
\label{tab:5}
\centering
\caption{Runtime performances of the GA searches for each shelter}

\begin{tabular}{cccccc}
\hline
\textbf{Shelter} & \textbf{Scenario} & \textbf{Runtime} & \textbf{\# Iteration} & \textbf{Resolution} & \textbf{\# AC} \\
\hline
A & - & 21 mins & 200 & 1 m & 212 \\
B / C & - & 26 mins & 200 & 1 m & 257 \\
D & I & 119 mins & 100 & 1.4 m & 253 \\
D & II & 140 mins & 100 & 1.4 m & 266 \\
\hline
\end{tabular}
\end{table}

These results demonstrate the efficiency and effectiveness of the proposed framework in optimizing the cage layouts for different shelters under various scenarios. The rapid convergence of the GA and the practical runtime performances highlight the applicability of this approach in real-world settings.

\section{Conclusion and Future Opportunities}
\label{sec:4}

Overpopulation and high noise levels are among the most critical issues adversely affecting the welfare of animal shelters. While numerous studies have proposed various measures and solutions to improve shelter environments, cage layout optimization has not been sufficiently addressed. To fill this gap, this study devises a multi-criteria layout optimization framework to generate optimal layouts based on management priorities.

This framework provides shelter management with a decision-making tool to systematically reduce overcrowding and tension in dog kennels while considering aisles and accessibilities. The study develops a graph-based algorithm to calculate the accessibility of the cages and the shortest paths to access points. Additionally, a heuristic algorithm analyzes the intensity of animal confrontations in the kennels. A TOPSIS technique is employed within a Genetic Algorithm (GA) to rank the stochastically proposed layouts at each iteration.

The results from three cats' kennel areas and two dogs' kennel areas demonstrate the framework's practicability. The MCDM characteristics of the framework make it adjustable to find solutions for different scenarios and requested capacities. The paper applies the framework to optimally place cages inside four shelters for cats and dogs in an animal care center plan, demonstrating the feasibility of the framework to find solutions within relatively short runtimes.

Several research opportunities and working directions are identified and recommended as follows: (1) Proposing more criteria for the layout evaluation phase to make the layouts compatible with some specific desires. (2) Employing different cage sizes inside each kennel area. (3) Considering training areas and human-animal interaction areas inside the layouts.


\bibliographystyle{unsrt}  
\bibliography{references}

\end{document}